\documentclass[conference,compsoc]{IEEEtran}

\ifCLASSOPTIONcompsoc
  \usepackage[nocompress]{cite}
\else
  \usepackage{cite}
\fi

\ifCLASSINFOpdf
  \usepackage[pdftex]{graphicx}
\else

\fi

\usepackage{svg}
\usepackage{multirow}
\usepackage{amsmath}
\usepackage{float}
\usepackage{flushend}

\begin{document}

\title{Fine-Tuning BERT for Sentiment Analysis of Vietnamese Reviews}

% \author{\IEEEauthorblockN{Quoc Thai Nguyen, Thoai Linh Nguyen }
%     \IEEEauthorblockA{University of Information Technology, Ho Chi Minh City, Vietnam \\
%     Vietnam National University, Ho Chi Minh City, Vietnam\\
%     Ho Chi Minh City, Vietnam\\
%     \{16521089,16521716\}@gm.uit.edu.vn}
% \and
%     \IEEEauthorblockN{Ngoc Hoang Luong, Quoc Hung Ngo}
%     \IEEEauthorblockA{University of Information Technology\\
%     Vietnam National University - HCM City\\
%     Ho Chi Minh City, Viet Nam\\
%     \{hoangln,hungnq\}@uit.edu.vn}
% }
\author{\IEEEauthorblockN{Quoc Thai Nguyen\IEEEauthorrefmark{1}\IEEEauthorrefmark{2},
Thoai Linh Nguyen\IEEEauthorrefmark{1}\IEEEauthorrefmark{2},
Ngoc Hoang Luong\IEEEauthorrefmark{1}\IEEEauthorrefmark{2}, and
Quoc Hung Ngo\IEEEauthorrefmark{1}\IEEEauthorrefmark{2}}
\IEEEauthorblockA{\IEEEauthorrefmark{1}University of Information Technology, Ho Chi Minh City, Vietnam}
\IEEEauthorblockA{\IEEEauthorrefmark{2}Vietnam National University, Ho Chi Minh City, Vietnam}
}
\maketitle
%-------------------------------------------------------------------------------
\begin{abstract}
Sentiment analysis is an important task    in     the   field of Nature Language
Processing (NLP), in which users' feedback data on a specific issue are evaluated 
and analyzed. Many deep learning models     have been proposed    to tackle this 
task, including the recently-introduced Bidirectional Encoder    Representations  
from  Transformers (BERT) model.  In this paper, we  experiment    with two BERT 
fine-tuning  methods for the sentiment analysis task on datasets of  Vietnamese 
reviews: 1) a method that uses only the [CLS] token as the input for an attached
feed-forward neural network, and 2) another method in      which all BERT output 
vectors are used as the input for classification. Experimental results     on two 
datasets show  that models using BERT slightly outperform other models  using 
GloVe and FastText.     Also, regarding the datasets employed in this study, our 
proposed   BERT fine-tuning method produces a model with better performance than 
the original BERT fine-tuning method.

\end{abstract}


\IEEEpeerreviewmaketitle

%-------------------------------------------------------------------------------
\section{Introduction} \label{intro}

Nowadays, customers often leave their reviews about various kinds of products and 
services, which they have used on a wide range of online platforms, from  
shopping    sites to social media. Such user experience information is a valuable 
source of reference for potential customers to make their buying decisions,   and 
especially, for related companies to improve their businesses. 
However,     it is prohibitively cumbersome to manually process such large amounts 
of data. Automated sentiment analysis systems are, therefore, essential to study 
customers' subjective opinions from their reviews in an efficient manner.
A fundamental task    of such systems is sentiment analysis, in which it is 
inferred  from the contents of each review whether the reviewer likes or dislikes 
the thing being discussed about.

In 2018,       Devlin et al. \cite{devlin2018bert} introduces a new language
representation model, namely BERT (Bidirectional Encoder Representations     from
Transformers).       This model has successfully improved recent works in finding 
representations of words in a digital space from their context. In this paper, we 
will present two methods for the sentiment analysis of Vietnamese reviews, and both 
are based on the BERT language model. Firstly,  we implement the   classification
method proposed in Jacob Devlin's study \cite{devlin2018bert}. Secondly, we propose  a new 
method for text classification that  integrates BERT into three   classification models. 
 We aim to exhibit the effectiveness of BERT for the sentiment analysis task compared 
to other models. We carry out other fine-tuning BERT methods for sentiment
analysis and select the best model to combine with BERT.

The next section   reviews related works for the sentiment analysis task.  Section 
\ref{sec:Background} gives  an overview of Word Embedding, Language model and BERT 
model. Section \ref{method} describes details of two methods for fine-tuning BERT on   
sentiment analysis.  The experimental results and analysis with our classification
approach are described in Section \ref{exper}. Finally, we conclude the  paper and  
give  some  future research directions in Section \ref{sec:Conclusion}.

%-------------------------------------------------------------------------------
\section{Related Work} 
\label{sec:Relatedwork}

Sentiment analysis is one of text classification tasks,  in which the input text 
is    classified  as  sentiment labels, such as \textit{“positive”} or
\textit{“negative”} label. There are several studies on this task with machine
learning approaches and lexicon/dictionary approaches (as shown in Table 
\ref{tab:PrevApproaches}). In 2002, the first research to classify the  reviews 
into two groups, \textit{positive} and \textit{negative} \cite{zou2015sentiment}. 
Several   studies     used supervised learning models to solve the task, such as, 
Support Vector Machine (SVM) \cite{Jadav2016SentimentAU}, 
\cite{Zainuddin2014SentimentAU},    Naïve Bayes \cite{Dey2016SentimentAO}, 
\cite{kavya2019sentiment}. In another study, they use a dictionary of emotional
vocabulary, which indicates whether a word is \textit{positive} or \textit{negative} 
along with it level \cite{taboada-etal-2011-lexicon}.

Recently, deep learning approaches allow to learn better representations for words 
based on their context. Yoon Kim used    Convolution Neural Network (CNN)
\cite{zhang2015characterlevel}  for the document classification task, the author 
process with each character-based. Mikolov et al.     \cite{le2014distributed}  
proposed    an unsupervised learning algorithm neural network   called “Paragraph 
vector”, which  is similar to the work in  \cite{mikolov2013distributed}. However, they 
not   using bag-of-words of context words as input, but an additional 
matrix of text with each column of matrix is a \textit{“Paragraph vector”}.

In Vietnamese text,  Duyen et al. use Naïve Bayes, Max Entropy Model and SVM  to 
classify reviews on the Agoda site \cite{Duyen2014AnES}, which  allows users to book
hotel rooms on travel occasions.  The results show that  the SVM model  achieves
the best result. Quan     et al. \cite{Vo2017MultichannelLM} employs a deep learning approach, proposing a model      that combines Long 
Short-Term Memory (LSTM) and CNN, namely        multi-channel LSTM-CNN for   
Vietnamese sentiment analysis.  The combined model achieves a performance better than both CNN and
LSTM alone. This approach is similar to \cite{Vo2019HandlingNM},  which    a deep learning model  is used for
managing negative comments on social networks. Word vectors are passed through the
CNN component, and the output is then       used as the input for   the LSTM network to perform 
classification.

\begin{table}[H]
    \centering
    \caption{Approaches for Sentiment Analysis}
    \label{tab:PrevApproaches}
    \resizebox{8cm}{!}{
    \begin{tabular}{|c|l|c|}
        \hline
             & \textbf{Description} & \textbf{Author}\\
        \hline
            \multicolumn{3}{|l|}{\textbf{English}}\\
        \hline
            Lexicon-based & Dataset positive/negative  & \cite{taboada-etal-2011-lexicon}\\
            Naïve Bayes & Dataset positive/negative & \cite{kavya2019sentiment} \cite{Dey2016SentimentAO}\\
            SVM &Dataset positive/negative& \cite{Jadav2016SentimentAU} 
             \cite{Zainuddin2014SentimentAU} \\
            TextCNN & This model is tested 8 Dataset &\cite{zhang2015characterlevel}\\
        \hline
            \multicolumn{3}{|l|}{\textbf{Vietnamese}} \\
        \hline
            Naïve Bayes& Positive/negative, 73.65\% F1-score   & \cite{Duyen2014AnES} \\
            MEM & Positive/negative, 82.6\% F1-score   & \cite{Duyen2014AnES} \\
            SVM & Positive/negative, 80.4\% F1-score   & \cite{Duyen2014AnES} \\
            CNN-LSTM & Positive/negative/neutral, 87.2\%, 59.6 F1-score& \cite{Vo2017MultichannelLM} \\
            CNN-LSTM & Positive/negative/neutral, 85.5\%, 72\% F1-score  & \cite{Vo2019HandlingNM} \\
        \hline
    \end{tabular}
    }
\end{table}

In summary, deep learning is currently the state-of-the-art approach for the field for sentiment 
analysis. However, Vietnamese resources for NLP tasks are limited, while popular word representation models, such as  Word2Vec or   GloVe, are fixed and not contextually flexible. In this study, we experiment with fine-tuning BERT 
to address the sentiment analysis task for Vietnamese language.

%-------------------------------------------------------------------------------
\section{Background} 
\label{sec:Background}

\subsection{Word Embedding}

Word Embedding    is a method of mapping each word into a multi-dimensional real
space, however its size is much smaller than the dictionary size. Several studies
introduced approaches for the word embedding step in NLP tasks. Tomas  Mikolos 
proposed Word2Vec \cite{mikolov2013distributed}, which is a statistical method  
for efficiently learning a standalone word embedding  from a text corpus. The
global Vectors for Word Representation (GloVe) algorithm is another method for 
efficiently learning word vectors, developed by Pennington et al. 
\cite{Pennington2014GloveGV}. Unlike context-free techniques (e.g., GloVe, word2vec), Bidirectional Encoder Representation from
Transformer (BERT) is a state-of-the-art language model that generates a contextual representation for each word, taking into account its neighboring words \cite{devlin2018bert}.

\subsection{Language Model}

The language model is a probability distribution over text sets.    The language
model can show how much probability a sentence          (or phrase) belongs to a
language. Language models analyze bodies of text     data to provide a basis for
their word predictions. 

\begin{center}
    $P(x_{1}….x_{n})$
\end{center}
where $x_{1}$,$x_{2}$,\dots,$x_{n}$  is the sequence    of words that make up a 
sentence, and $n$ is length of the sentence (n $>$ 1).

%-------------------------------------------------------------------------------
Currently,    pre-trained language    models are widely used in NLP tasks.   These
models have been trained on a very large dataset, integrating many languages and
knowledge. Users     can thus use             them in many different NLP
tasks \cite{devlin2018bert}.

\subsection{BERT} 

BERT is a multi-layered structure of   Bidirectional Transformer encoder layers,
based on the architecture of transformer \cite{vaswani2017attention}.  BERT uses 
Bidirectional    Transformer encoders replace       Encoders combining Decoders.
BERT fine-tuning    tasks do not require Decoder blocks. Therefore, they replace 
Decoder blocks with similar Encoder blocks. 

The most    advantage of BERT is to    apply    two-dimensional training techniques
of Transformers from a  very famous Attention model to a Language Model.    Different from previous NLP studies, which      look at a text string from
left to right, BERT combines how to look at a text string from  2 dimensions
(from left to right and right to left).  This method         can greatly improve 
the retention of Representations of words in sentences.    The published results
also show that the trained language model has a more profound contextual meaning
than previous models.

%image1

Two sizes have been proposed for the BERT model: 
\begin{itemize}
    \item $BERT_{BASE}$: 12 encoder block,     12 head attention,    110 million parameters.
    \item $BERT_{LARGE}$: 24 encoder block,      16 head attention, 340  million parameters.
\end{itemize}

%-------------------------------------------------------------------------------
\section{Methodology} \label{method}

For the sentiment analysis task of review texts, BERT can be used with two approaches:
\begin{itemize}
    \item \textbf{Feature extraction:}      This method   uses BERT as a feature 
    extraction model. The architecture of the BERT model is preserved,  
    and its outputs are feature input vectors for subsequent classification models to solve the given problem.
    \item \textbf{Fine-tuning: }In this method, we need to modify 
    the architecture of the model by adding some layers at the end of BERT
    model. These layers will solve the problem,    and retrain the
    model.    This process is called fine-tuning. In \cite{devlin2018bert}, this 
    method is also the main method used    to evaluate benchmarks on   different
    tasks, showing BERT superiority over previous models.
\end{itemize}

\begin{table}[H]
    \centering
    \caption{Comparison between Fine-tuning and Feature-based Approaches with BERT \cite{devlin2018bert}}
    \label{tab:ComparisonResult}
    \resizebox{8cm}{!}{
    \begin{tabular}{|c|c|c|}
        \hline
            \textbf{SYSTEM} & \textbf{Dev F1} & \textbf{Test F1} \\
        \hline
            \textbf{Fine-tuning approach} & {} & {}\\
            $BERT_{LARGE}$ & 96.6 & 92.8\\
            $BERT_{BASE}$ & 96.4 & 92.4\\
        \hline
            \textbf{Feature-based approach ($BERT_{BASE}$)} & {} & {}\\
            Embeddings & 91.0 & -\\
            Second-to-Last Hidden & 95.6 & -\\
            Last Hidden & 94.9 & -\\
            Weighted Sum Last Four Hidden & 95.9 & -\\
            Concat Last Four Hidden & 96.1 & -\\
            Weighted Sum All 12 Layers & 95.5 & -\\
        \hline
    \end{tabular}}
    \vspace{-2mm} %Put here to reduce too much white space after your table 
\end{table}
\setlength{\textfloatsep}{0.1cm}
%-------------------------------------------------------------------------------
Table \ref{tab:ComparisonResult} (extracted from  \cite{devlin2018bert}) shows the performance of feature-based and fine-tuning approaches with BERT on the Named Entity Recognition task(CoNLL-2003 NER \cite{sang2003introduction}). Fine-tuning approaches achieve better
performance than feature-based approaches, where different combinations of hidden vectors are experimented with. 
Therefore, in this paper, we decide  to use the fine-tuning approach with BERT for our sentiment analysis task of Vietnamese reviews.

To fine-tune BERT,    we need to employ a pre-trained BERT model. It is required that the pre-trained model has been trained with Vietnamese datasets. In this work, we
use the pre-trained BERT-Base Multilingual 
Cased\footnote{https://github.com/google-research/bert}        provided by
Google because it supports multiple
languages,      including Vietnamese along with other languages (104 languages).
This pre-trained model is thus suitable for Vietnamese. Two methods of fine-tuning BERT for sentiment analysis are as follows.

\begin{figure}[ht]
    \centering
    \includegraphics[scale=0.7]{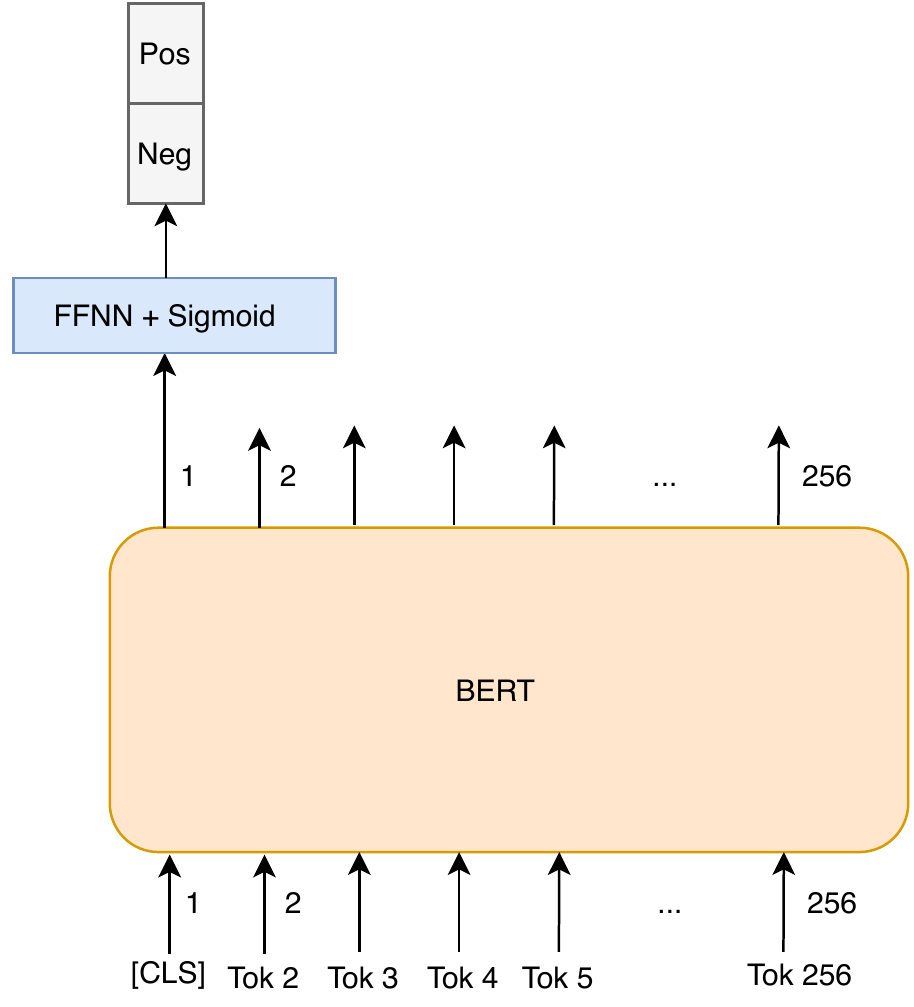}
    \caption{BERT-base architecture }
    \label{fig:Cover2}
\end{figure}

\begin{figure}[ht]
    \centering
    \includegraphics[scale=0.6]{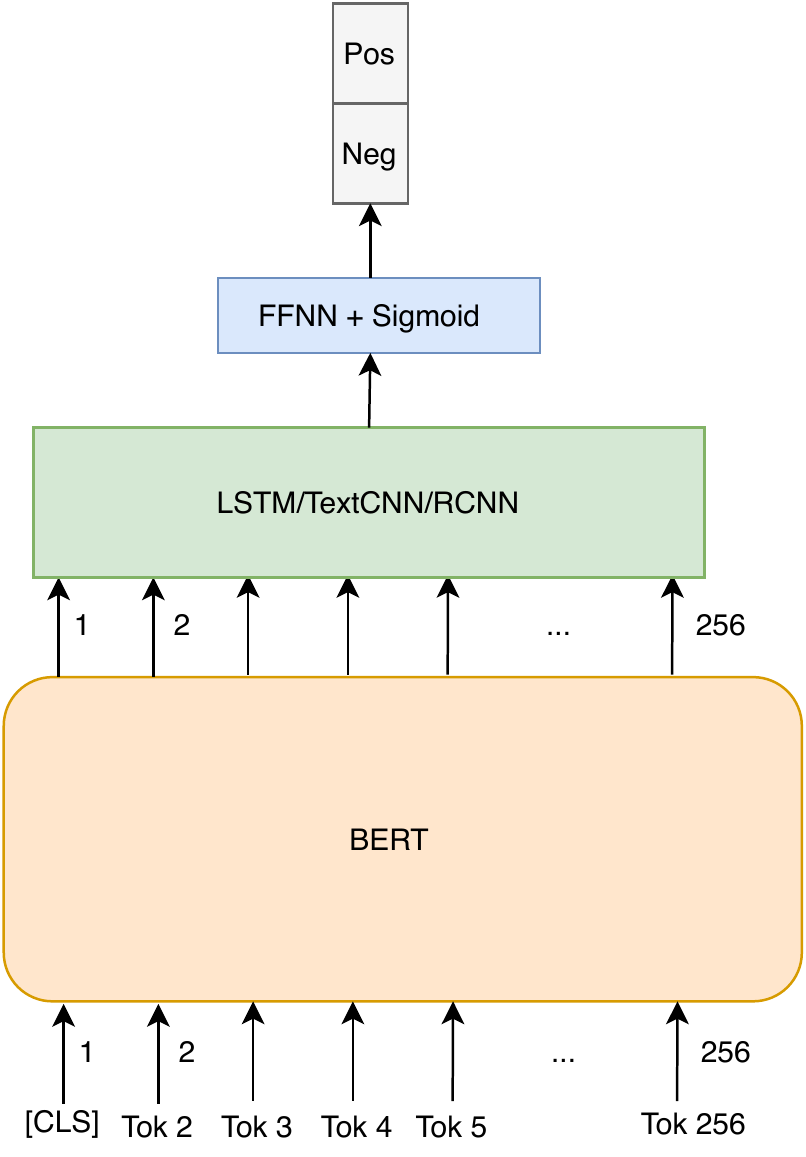}
    \caption{BERT architecture using all tokens}
    \label{fig:Cover3}
\end{figure}

\begin{itemize}
    \item \textbf{Fine-tuning BERT using token [CLS]:}
    Devlin et al. \cite{devlin2018bert} perform   fine-tuning of BERT at the sequence level. They 
    add a special token to perform the classification  tasks. A token [CLS]   is
    added to the beginning position of the sentences.  The output vector of this
    token will be sent through the feedforward neutral network to perform input
    sentence classification (as shown in Figure \ref{fig:Cover2}). This model is 
    named  \textit{BERT-base}.
    \item \textbf{Fine-tuning BERT using all tokens:} We  use the entire output
    of BERT including the token [CLS].       These outputs form a $SEQ\_LEN \times h$ matrix, where $SEQ\_LEN$ is the maximum length of the
    input sequence, and   $h$ is the length of hidden vectors. We can then use this
    output matrix as the input to other classification model which can be one of  
    three models: LSTM, TextCNN, or RCNN (as shown in Figure \ref{fig:Cover3}).
\end{itemize}

    %image2(bert_base)

\textbf{Long Short-Term Memory (LSTM):} We employ the Long Short-Term Memory (LSTM), which is the most widely-used recurrent neural network (RNN) model, that addresses the distance dependence issue of the classical RNN model. The LSTM layer then continues to extract the features received from BERT.

\textbf{Text Convolution Neural Network (TextCNN):} The convolution model in this research  is TextCNN \cite{kim2014convolutional}.    This is the CNN model widely used in nature language processing tasks,  especially for classification tasks thanks  to the quick training time and good results.     The TextCNN model employed in this work
is a little         different from the      TextCNN models       suggested      in \cite{kim2014convolutional, zhang2015sensitivity}. We use 4 regions sizes (2,3,4,5).
%and only use 1 filter for each region size.

%image5(TextCNN)
\begin{figure}[h]
    \centering
    \includegraphics[scale=0.35]{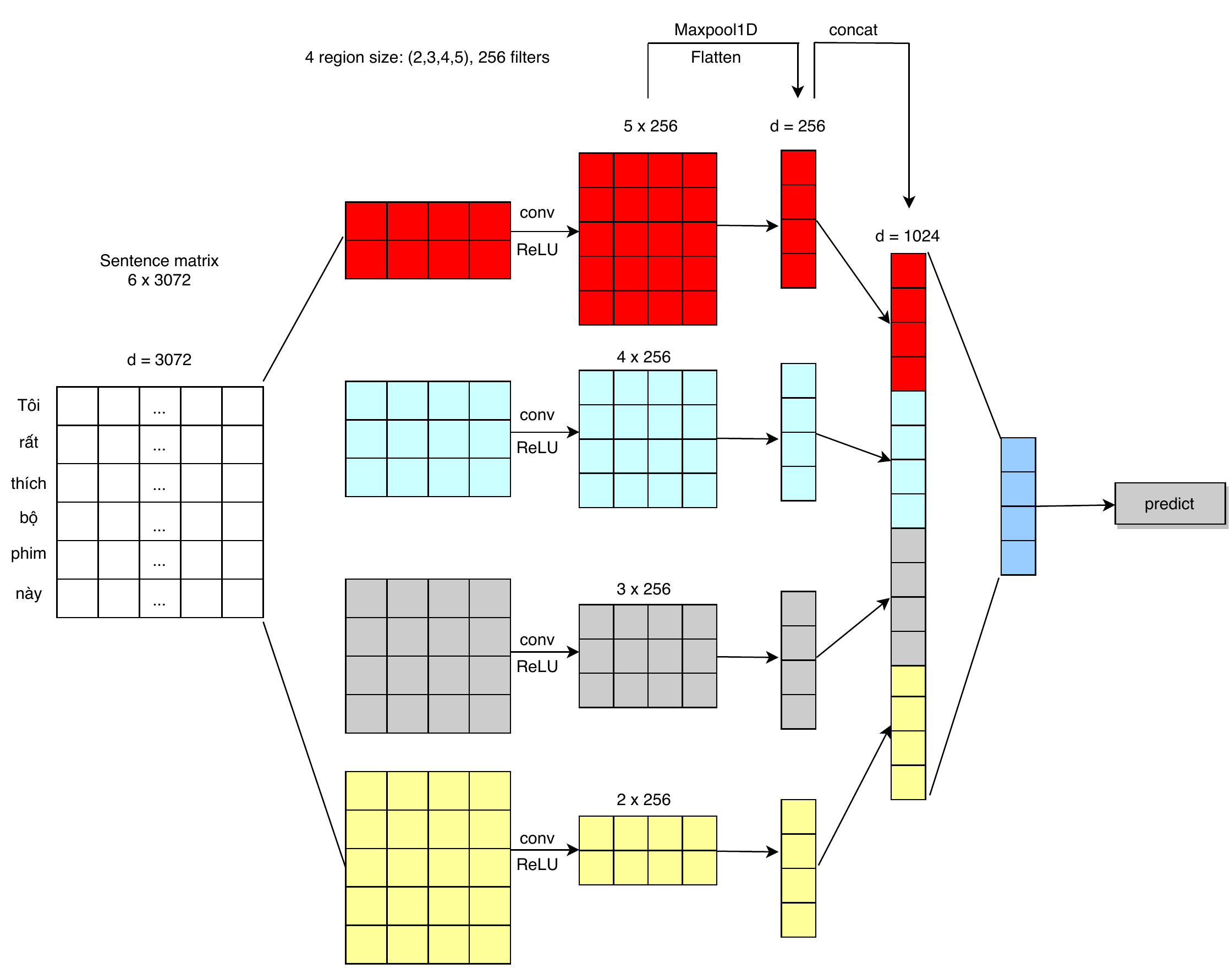}
    \caption{TextCNN architecture \cite{zhang2015sensitivity}}
    \label{fig:Cover5}
\end{figure}

\begin{figure*}[t]
    \centering
    \includegraphics[scale=0.7]{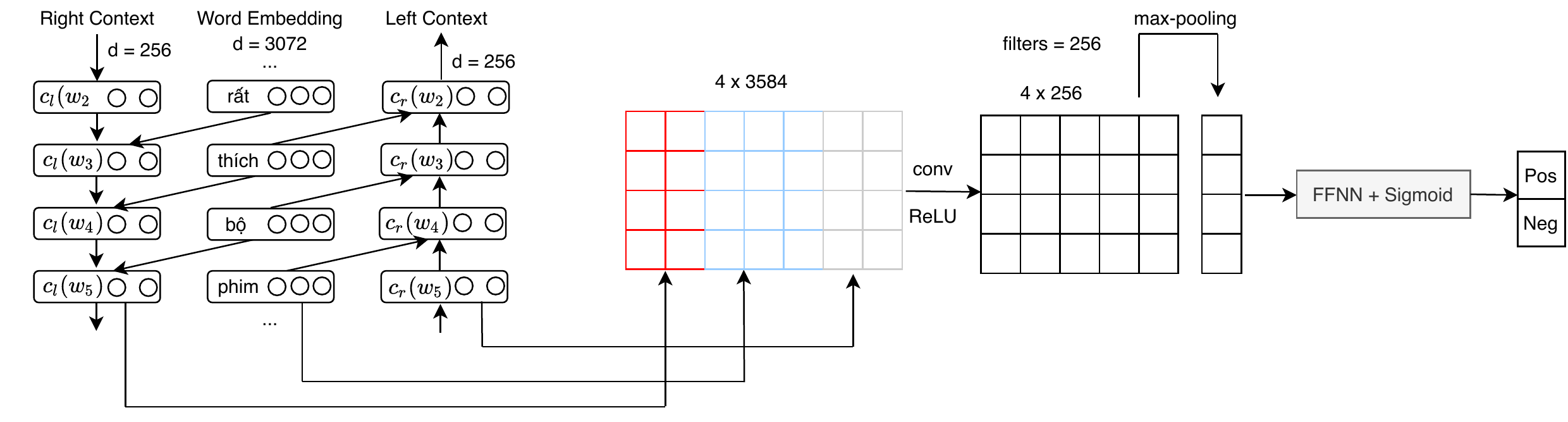}
    \caption{RCNN Architecture}
    \label{fig:Cover6}
\end{figure*}

\textbf{Recurrent Convolutional Network (RCNN)}: The model   is a combination of 
recurrent and convolutional architectures \cite{Lai2015RecurrentCN}. Two LSTM architectures are combined:    one earns the context of
words from left to right while the other learns contextually from right to left.
The  output of both networks is then passed through a \textit{conv1d} layer to continue extracting. The RCNN architecture is described in Figure \ref{fig:Cover6}.
%image6(RCNN)

%Figure  \ref{fig:Cover4},    
Figure \ref{fig:Cover5},    Figure \ref{fig:Cover6}
illustrate the architectures that will be  associated with BERT in this work.
To perform       classification   of \textit{Positive} and \textit{Negative} labels,    we use the logistics function.%that is $Sigmoid$.

Based on the experimental results of \cite{devlin2018bert}, fine-tuning approaches give better performance than  
feature-based approaches (as shown in Table \ref{tab:ComparisonResult}). Among the feature-based approaches, the one that concatenate the last 
four hidden outputs gives the best results. In this paper, we decide to also concatenate the last four hidden outputs when performing fine-tuning.
Each hidden output has $dim = h$. Therefore, concatenating four hidden outputs results in the dimensions of the output  vector having $dim = 4 \times  h.$

%-------------------------------------------------------------------------------
\section{Experiments \& Discussions} \label{exper}
\subsection{Datasets}

We employ two datasets to train and evaluate all the models that are investigated in this paper.
The two datasets consist of reviewing texts commented by users on e-commerce sites.   

First, the \textit{Ntc-sv}\footnote{https://streetcodevn.com/blog/sav}         dataset
includes reviews about food and restaurants on Foody.  This dataset consists of
50,000 samples.         The labels      are assigned based on the average score (avg\_score): the ones above 8.5 are labeled \textit{positive},     the ones less than 5 are labeled \textit{negative}.

Second, the \textit{Vreview} dataset is built on the training dataset of a competition of \textit{AIVIVN}\footnote{https://www.aivivn.com/contests/1}   on sentiment analysis, which consists users' comments of product reviews on various e-commerce sites.    In addition, we include some comments      about        food and restaurants on 
\textit{Foody}\footnote{https://forum.machinelearningcoban.com/t/du-lieu-review-cua-foody/203}.
We then perform data labeling through the average score
(avg\_score) similarly to the Ntc-sv  dataset. 
However,  a slightly different thresholds are used here: the samples with average scores above 7.5 are labeled \textit{positive}, and the ones  below 5 are labeled \textit{negative}.

Employing two datasets with different characteristics would help to better evaluate the performance of different models.
Some data statistics are made for the evaluation        process are showed in Tables \ref{tab:table2} and \ref{tab:table3}. These are the statistics after the data have been processed.

\begin{table}[ht]
    \centering
    \caption{Data Description after Pre-process}
    \label{tab:table2}
    \resizebox{8cm}{!}{
    \begin{tabular}{|c|c|c|c|c|c|}
        \hline
            \multirow{2}{*}{\textbf{Dataset}} & 
            \multicolumn{2}{|c|}{\text{\textbf{Train}}}&
            \multicolumn{2}{|c|}{\text{\textbf{Test}}} &
            \multirow{2}{*}{\textbf{Totally}} \\
            \cline{2-5}
            {} & Positive & Negative & Positive & Negative & {}\\
        \hline
            \textbf{Ntc-sv} & 20,493 & 20,267 & 5,000 & 5,000 & 50,760 \\
        \hline
            \textbf{Vreview} & 22,979 & 19,537 & 8,301 & 6,795 & 57,612 \\
        \hline
    \end{tabular}}
    
\end{table}

\begin{table}[h]
    \centering
    \caption{Statistics of words included in the comments}
    \label{tab:table3}
    \resizebox{4cm}{!}{
    \begin{tabular}{|c|c|c|}
        \hline
            {} & \textbf{Vreview} & \textbf{Ntc-sv} \\
        \hline
            \textbf{Mean} & 55.45 & 86.57\\
        \hline
            \textbf{Stdn} & 63.75 & 77.41\\
        \hline
            \textbf{Min} & 1 & 1\\
        \hline
            \textbf{25\%} & 14 & 37\\
        \hline
            \textbf{50\%} & 32 & 65\\
        \hline
            \textbf{75\%} & 76 & 111\\
        \hline
            \textbf{Max} & 435 & 1,501\\
        \hline
    \end{tabular}}
    
\end{table}

\subsection{Models}
In this paper, we employ the following machine learning and deep learning models in comparison with the BERT models for the sentiment analysis task over our datasets of Vietnamese reviews.
\begin{itemize}
    \item \textbf{SVM / Boosting:} SVM and Boosting are two classic        machine learning algorithms. In this study, SVM experiments are based on $n$-grams features     ($n$ in the range [1,5]),  while Boosting experiments are 
    based on the XGBoosting algorithm \cite{Chen2016XGBoostAS} (deep = 15).
    \item \textbf{FastText + LSTM / TextCNN / RCNN:} We choose  three deep learning architectures 
    to combine with  the word embedding model FastText \footnote{https://fasttext.cc/docs/en/crawl-vectors.html}, which was
     pre-trained on a Vietnamese dataset.
    \item\textbf{GloVE + LSTM / TextCNN / RCNN:} Because we do not
    have pre-trained GloVE embeddings with the same dimensions as     the pre-trained FastText embedding, we use the glove-python\footnote{https://github.com/maciejkula/glove-python}
    library to train the word embedding model. Models associated with
    GloVE would then be similar to models combined with FastText, ensuring fair comparisons in our experiments.
\end{itemize}

\subsection{Results \& Discussions}

The performance results of all the competing models on the two datasets are shown in Tables \ref{tab:table4} and \ref{tab:table5}.

\begin{table}[ht]
    \centering
    \caption {Experimental results on ntc-sv dataset}
    \label{tab:table4}
    \resizebox{8cm}{!}{
    \begin{tabular}{|c|c|c|c|}
        \hline
            \textbf{Model} & \textbf{Precision(\%)} & 
            \textbf{Recall(\%)} & \textbf{F1(\%)}\\
        \hline
            SVM &	89.23 & 92.52 & 90.84\\
        %\hline
            XGBoost & 88.76 & 90.58 & 89.63\\
        \hline
            FastText + TextCNN & 67.9 & 89.1 & 77.1\\
        %\hline
            FastText + LSTM & 88.5 & 89.7 &	89.1\\
        %\hline
            FastText + RCNN & 89.2 & 91.7 & 90.4\\
        \hline
            Glove + TextCNN & 69.7 & 87.7 & 77.7\\
        %\hline
            Glove + LSTM & 88.7 & 91.8 & 89.8\\
        %\hline
            Glove + RCNN & 85.8 & 85.8 & 90.7\\
        \hline
            BERT-base &	88.13 &	\textbf{94.02} & 90.9\\
        %\hline
            BERT-LSTM & \textbf{89.78} & 92.08 & 90.91\\
        %\hline
            BERT-TextCNN & 88.85 & 93.14 & 90.94\\
        %\hline
            BERT-RCNN & 88.76 & 93.68 & \textbf{91.15}\\
        \hline
    \end{tabular}}
    
\end{table}

\begin{table}[ht]
    \centering
    \caption {Experimental results on vreview dataset}
    \label{tab:table5}
    \resizebox{8cm}{!}{
    \begin{tabular}{|c|c|c|c|}
        \hline
            \textbf{Model} & \textbf{Precision(\%)} & \textbf{Recall(\%)} & \textbf{F1(\%)}\\
        \hline
            SVM &	86.26 & 86.9 & 86.5\\
        %\hline
            XGBoost & 87.69 & 88.45 & 88.07\\
        \hline
            FastText + TextCNN & 61.8 & \textbf{94} & 74.6\\
        %\hline
            FastText + LSTM & 88.5 & 86.4 &	87.5\\
        %\hline
            FastText + RCNN & 84.5 & 89.8 & 87.1\\
        \hline
            Glove + TextCNN & 62.6 & 93 & 74.8\\
        %\hline
            Glove + LSTM & 85.8 & 85.8 & 85.8\\
        %\hline
            Glove + RCNN & 84.0 & 88.6 & 86.2\\
        \hline
            BERT-base &	86.08 &	88.44 &	87.2\\
        %\hline
            BERT-LSTM & 85.25 & 89.9 & 87.5\\
        %\hline
            BERT-TextCNN & \textbf{90.9} & 85.2 & 87.98\\
        %\hline
            BERT-RCNN & 87.08 & 89.38 & \textbf{88.22}\\
        \hline
    \end{tabular}} 
    
\end{table}

The following conclusions can be drawn from the experiments with BERT-based approaches and other approaches:
\begin{itemize}
    \item When comparing the models on  the ntc-sv dataset, it can be seen that the
    results are not considerably different,  even when comparing BERT models with SVM.
    Therefore, deep learning  models are not much superior in this case. The 
    training time of BERT-based approaches is  much longer than the training time 
    of the SVM approach; however, their performances are not much better than SVM's.
    \item Models using the word embeddings FastText    and    GloVe produce similar
    results. First, TextCNN models produce very poor results,    much lower than
    other models. Secondly,   the models using architectures of LSTM and RCNN,
    especially models using RCNN architecture produce very good results, because 
    it combines the features that are received from many sources,  from which there is
    more information available for better classifying the reviews.  Their training times, however, are considerably longer.
    \item The BERT-based, BERT-LSTM, and BERT-TextCNN approaches similarly exhibit good performance on both datasets.
    \item BERT-RCNN is the model that gives the best results among the models considered in this paper. It can be seen that RCNN is not only suitable when combined with BERT but also effective in combination with other embedding methods. The experimental results here suggest that RCNN is the most suitable approach to be combined with BERT to implement fine-tuning.
    %in the second way with the very good results that it brings.
\end{itemize}

The experimental results did not show a considerable difference between BERT and other models, although we used a different fine-tuning approach, which made the most of the information from BERT. These results can be explained with the following reasons:

\begin{itemize}
    \item The lengths of the comments are relatively short  and the pre-processing procedure eliminates most of the potential  confusions, making it easier to classify the comments.
    \item All the fine-tuning BERT approaches, BERT-base and BERT in combination with other deep learning architecture, require a lot of computational resources,  namely RAM and GPU. The input
    string for BERT after sub-splitting is 512, but our resources are not
    enough to be able to fine-tune BERT with the input string length of 512. If we
    had enough resources to process the maximum  input length, 
    the performance of BERT-based methods could be further improved.
    \item Pre-trained BERT-Base Multilingual Cased     might still be sub-optimal for Vietnamese language.  The
    sub-word separator in     BERT uses the sub-word separation    technique for
    English, which might lead to inaccurate separations when being applied to Vietnamese.   The
    pre-trained BERT model, therefore, has certain contextual deviations for Vietnamese words. 
\end{itemize}

The source code, models and datasets of this study can be downloaded at: {\color{blue} https://github.com/thoailinh/Sentiment-Analysis-using-BERT}.

%-------------------------------------------------------------------------------
\section{Conclusion} 
\label{sec:Conclusion}

We have fine-tuned BERT using the pre-trained multilingual   BERT with two approaches  and both exhibit better performance  compared to other machine learning and deep learning models considered in this paper.
When performing fine-tuning BERT, the entire output of BERT can be employed, providing the subsequent classification model with more useful information.
The experimental results showed that fine-tuning BERT with LSTM or TextCNN yielded no considerable improvement compared to the BERT-base fine-tuning approach. 
Instead, BERT could be integrated with models using RCNN or other architectures that combine recurrent and convolutional models.

The experimental results show that the BERT-RCNN model using our proposed fine-tuning method yields certain improvement of the accuracy performance for sentiment analysis on datasets containing Vietnamese reviews. Although the improvement is not considerably superior, it exhibits potentials for further enhancements. For future work, we aim to extend the proposed method for aspect-based sentiment analysis, and more extensive experiments could be conducted on different datasets.

%-------------------------------------------------------------------------------


\begin{thebibliography}{1}

\bibitem{Dey2016SentimentAO}
Lopamudra Dey Sanjay Chakraborty et al. , ``Sentiment Analysis of Review Datasets 
Using Naïve Bayes and          K-NN Classifier",    in International Journal of 
Information Engineering and Electronic Business , 2016.

\bibitem{vaswani2017attention}
Ashish Vaswani	et al. ``Attention is all you need”, 31st Conference on Neural
Information Processing Systems (NIPS 2017), Long Beach, CA, USA.

\bibitem{zou2015sentiment}
Bo Pang, Lillian Lee, and Shivakumar Vaithyanathan,    ``Sentiment classification 
using machine learning techniques",    in Proceedings of Conference on Empirical 
Methods in Natural Language Processing (EMNLP-2002), 2002.

\bibitem{sang2003introduction}
Erik F Tjong Kim Sang and Fien De Meulder,  ``Introduction to the conll-2003 
shared task: Language-independent named entity recognition". In CoNLL (2003).

\bibitem{devlin2018bert}
Jacob Devlin, et al. ``BERT: Pre-training of Deep Bidirectional Transformers  for
Language Understanding”. 2018.

\bibitem{Jadav2016SentimentAU}
Bhumika M. Jadav and Vimalkumar B. Vaghela,  ``Sentiment Analysis using Support 
Vector Machine based on Feature Selection and Semantic Analysis", in International 
Journal of Computer Applications 2016.

\bibitem{Zainuddin2014SentimentAU}
Nurulhuda Zainuddin and Ali Selamat, ``Sentiment analysis using Support Vector 
Machine", 2014 International Conference on Computer, Communications, and Control 
Technology (I4CT).

\bibitem{Pennington2014GloveGV}
Jeffrey Pennington, Richard Socher, Christopher Manning, ``GloVe: Global Vectors 
for Word Representation”,    in Proceedings of the 2014 Conference on Empirical 
Methods in Natural Language Processing (EMNLP), 2014.

\bibitem{taboada-etal-2011-lexicon}
Maite Taboada, Julian Brooke, Milan Tofiloski, Kimberly Voll, and Manfred Stede,
``Lexicon-Based Methods for Sentiment Analysis", Computational Linguistics,  vol. 
37, no. 2, pp. 267-307, 2011.

\bibitem{Vo2019HandlingNM}
Khuong Vo, Tri Nguyen, Dang Pham, Mao Nguyen, Minh Truong, Dinh Nguyen, Tho Quan,
``Handling negative mentions on social media channels using deep learning", Journal 
of Information and Telecommunication, vol. 3, no. 3, 2019.

\bibitem{Duyen2014AnES}
Nguyen Thi Duyen,      Ngo Xuan Bach,      Tu Minh Phuong, ``An Empirical Study on 
Sentiment Analysis for Vietnamese”. The 2014 International Conference on Advanced
Technologies for Communications (ATC'14).

\bibitem{turney2002thumbs}
Peter D. Turney, ``Thumbs up or thumbs down?: semantic orientation applied     to
unsupervised classification of reviews", in Proceedings of Annual Meeting of the
Association for Computational Linguistics (ACL- 2002), 2002.

\bibitem{Lai2015RecurrentCN}
Siuwei Lai, Liheng Xu, Kang Liu, Jun Zhao, ``Recurrent Convolutional Neural Networks 
for Text Classification”,     Proceedings of the Twenty-Ninth AAAI Conference on 
Artificial Intelligence (2015).

\bibitem{kavya2019sentiment}
Kavya Suppala, Narasinga Rao, ``Sentiment Analysis Using Naïve Bayes Classifier", 
in International Journal of Innovative Technology and Exploring Engineering 
(IJITEE) 2019

\bibitem{Chen2016XGBoostAS}
Tianqi Chen, Carlos Guestrin, ``XGBoost: A Scalable Tree Boosting System”, The 
22nd ACM SIGKDD International Conference (2016).

\bibitem{zhang2015characterlevel}
Xiang Zhang, Junbo Zhao, Yann LeCun, ``Character-level Convolutional Networks for 
Text Classification", arXiv:1509.01626, 2015.

\bibitem{zhang2015sensitivity}
Ye Zhang, Byron C. Wallace, ``A Sensitivity Analysis of (and Practitioners’ Guide 
to) Convolutional Neural Networks for Sentence Classification”, IJCNLP 2017.

\bibitem{kim2014convolutional}
Yoon Kim ``Convolution Neutral Network for Sententce Classification”, Proceedings 
of   the 2014 Conference on Empirical Methods   in   Natural Language Processing 
(EMNLP).

\bibitem{le2014distributed}
Quoc Le, Tomas Mikolov, ``Distributed Representations of Sentences and Documents", 
in   Proceedings of the International Conference on Machine Learning (ICML 2014), 
2014.

\bibitem{Vo2017MultichannelLM}
Quan-Hoang Vo, Huy-Tien Nguyen, Bac Le, Minh-Le Nguyen, ``Multi-channel  LSTM-CNN 
model   for Vietnamese sentiment analysis”, 2017 9th International Conference on
Knowledge and Systems Engineering (KSE).

\bibitem{mikolov2013distributed}
Tomas Mikolov, Ilya Sutskever, Kai Chen, Greg Corrado,    and      Jeffrey Dean ,
``Distributed Representations of Words and Phrases and their Compositionality", in
Proceedings of the 26th International Conference on Neural Information Processing
Systems (NIPS 2013), 2013.

\end{thebibliography}
\end{document}